\documentclass{article}

\usepackage{arxiv}

\usepackage[utf8]{inputenc} 
\usepackage[T1]{fontenc}    
\usepackage{lmodern} 
\usepackage{url}            
\usepackage{booktabs}       
\usepackage{amsfonts}       
\usepackage{nicefrac}       
\usepackage{microtype}      
\usepackage{lipsum}         
\usepackage{graphicx}
\usepackage{caption}
\usepackage{setspace}
\usepackage{doi}
\usepackage{hyperref}
\hypersetup{
    linktocpage=true,
    colorlinks=true,
    linkcolor=blue,
    citecolor=blue,
    urlcolor=blue,
    }
\usepackage{cleveref}       

\title{An efficient neuromorphic approach\\for collision avoidance combining\\Stack-CNN with event cameras}

\date{}

\newif\ifuniqueAffiliation

\ifuniqueAffiliation 
\author{ 
    {\includegraphics[scale=0.12]{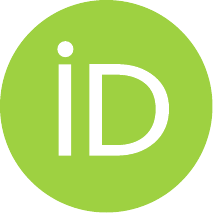}
    \hspace{1mm}Antonio Giulio Coretti} \\
	Department of Physics\\
	University of Turin\\
	Via Pietro Giuria 1,10125 Turin, Italy \\
	\texttt{hippo@cs.cranberry-lemon.edu} \\
	\And
	\href{https://orcid.org/0009-0003-7430-0639}{\includegraphics[scale=0.12]{orcid.pdf}
    \hspace{1mm}Mattia Varile} \\
	AIKO S.r.l.\\
	Mount-Sheikh University\\
	Santa Narimana, Levand \\
	\texttt{stariate@ee.mount-sheikh.edu} \\
	\And
	\href{https://orcid.org/0000-0000-0000-0000}{\includegraphics[scale=0.12]{orcid.pdf}
    \hspace{1mm}Elias D.~Striatum} \\
	Department of Electrical Engineering\\
	Mount-Sheikh University\\
	Santa Narimana, Levand \\
	\texttt{stariate@ee.mount-sheikh.edu} \\
}
\else
\usepackage{authblk}

\newbox{\orcid}\sbox{\orcid}{\includegraphics[scale=0.08]{orcid.pdf}} 
\author[1,2,3]{%
	\href{https://orcid.org/0009-0003-7430-0639}{
        \usebox{\orcid}\hspace{1mm}Antonio Giulio Coretti
        \thanks{Corresponding author, \texttt{antoniogiulio.coretti@unito.it}}}%
}
\author[3]{
	\href{https://orcid.org/0000-0001-9833-2537}{\usebox{\orcid}\hspace{1mm}Mattia Varile}
}
\author[1,2]{
	\href{https://orcid.org/0000-0003-1069-1397}{\usebox{\orcid}\hspace{1mm}Mario Edoardo Bertaina}
}

\affil[1]{Department of Physics, University of Turin, Via Pietro Giuria 1, 10125 Turin, Italy}
\affil[2]{INFN, Section of Turin, Via Pietro Giuria 1, 10125 Turin, Italy}
\affil[3]{AIKO S.r.l., Via dei Mille 22, 10123 Turin, Italy}
\fi

\renewcommand{\headeright}{  }
\renewcommand{\undertitle}{  }
\renewcommand{\shorttitle}{\small \flushleft Published as a Conference Paper at \textsc{SpaceOps2025}}

\hypersetup{
    pdftitle={An efficient neuromorphic approach for collision avoidance combining Stack-CNN with event cameras},
    pdfsubject={q-bio.NC, q-bio.QM},
    pdfauthor={Antonio Giulio Coretti},
    pdfkeywords={space debris, Space Situational Awareness, Space Traffic Management, collision avoidance, event camera, deep learning},
}

\begin{document}
\maketitle

\begin{abstract}
    Space debris (SD) represents a serious threat for all space activities. Several studies have been carried out to tackle the problem, some of these involving an active management of new SD coming from inactive satellites, parking them in graveyard orbits or deorbiting them until complete disintegration, while others suggesting the removal of resident SD using capture mechanisms. By contrast, another line of research addresses the problem in a passive way, accepting space junk and developing collision avoidance techniques. Fitting into the latter context of research, this work presents an innovative collision avoidance system making use of a recently developed type of imaging device, the event-based camera. These bio-inspired sensors operate in a significantly different manner with respect to the traditional frame-based cameras, making them well-suited not only for computer vision and robotics, but also for Space Situational Awareness (SSA) and Space Traffic Management (STM). In this work, the architecture of a collision avoidance system based on the real-time analysis of an event-based camera data stream using a deep learning algorithm is proposed. Called Stack-CNN and previously used for meteor detection, the algorithm makes use of a Stacking procedure according to a trial velocity vector, before applying a Convolutional Neural Network (CNN) to distinguish the signal from the background. The algorithm has been tested on SD data recorded by terrestrial event-based cameras, demonstrating great results in detecting faint moving objects due to its ability to enhance the signal-to-noise ratio of the data stream. With the future possibility of coupling this type of neuromorphic sensors to recently developed neuromorphic processors, further reducing the absorbed power, this innovative type of camera might be extremely successful in the on-board space imaging field, substantially impacting space operations related to STM and SSA.
\end{abstract}


\section{Introduction}

\subsection{Background and motivation}

SD, also known as \textit{space junk}, constitutes a major challenge for the entire space community~\cite{001}. Mainly made up of not-removed spacecraft at the end of their service life, upper stages of lunch vehicles and fragments formed after collisions, SD represents the form of human pollution of the space environment. Since the launch in 1957 of the first satellite into space, Sputnik 1, there has been a continuous increase in the number of satellites and space missions leading to the rise of space debris, which year by year pose an ever-increasing risk for active spacecraft. Due to the continuous rise in the probability of future collisions, associated with the endless growth of its amount, studies on the topic are becoming more and more frequent in the scientific community. \\
In this context, collision avoidance is critical for ensuring the longevity and safety of space operations. If the SD is large enough (typically over $10\,\mbox{cm}$) to be tracked from Earth and cataloged, collision avoidance maneuvers can be notified to the orbiting satellite directly from ground stations well in advance. But there is a much more abundant population of SD, the one under 10 cm size, which constitutes the greatest hazard~\cite{002}. Indeed, in case of small-size SD, ground instruments usually fail the detection, and it is therefore necessary to have an on-board anti-collision system that allows not only the timely detection of small SD on a collision course, but also the prompt notification to the On-board Computer (OBC) in order to carry out a Collision Avoidance Maneuver (CAM). \\
Traditional collision avoidance systems often rely on a frame-based sensor for the detection of the incoming SD, coupled with a tracking algorithm to determine the risk of collision and, if necessary, to alert the OBC for carrying out the anti-collision maneuver. However, frame-based cameras may struggle with the dynamic nature of SD detection, leading researchers to the exploration of different types of sensors.

\subsection{Problem Statement}

Current collision avoidance systems face challenges in accurately detecting fast-moving, faint objects against the vastness of space. This paper addresses these issues by proposing a system that combines event-based cameras with Stack-CNN to improve the speed of the detection and its accuracy. The former is addressed through the very high spatial resolution and the very low latency associated with event cameras, features that enable a timely detection of SD. The latter, instead, is achieved thanks to the Stack-CNN algorithm's ability to increase the signal-to-noise ratio (SNR) of the scene, allowing extremely faint objects to be detected and consequently larger objects to be seen earlier. Moreover, the sparse nature of the data produced by the event-based camera allows to process the impending threat without computational overhead.

\subsection{Paper organization}

Section \ref{sec2} will give an overview of collision avoidance methods for SD and how they are currently addressed, followed by an analysis of the current use of event-based cameras in the fields of robotics and autonomous systems, and by a discussion of machine learning (ML) methods for object detection tasks in space. Then, Section \ref{sec3} will present the proposed system, characterizing its neuromorphic sensor and its Stack-CNN detection algorithm. Section \ref{sec4} will provide some preliminary results on the proposed coupling between sensor and deep learning detection method, while Section \ref{sec5} will discuss the possible limitations of the system and possible future studies. Finally, Section \ref{sec6} will summarize the content of the entire paper.

\section{Related Work} \label{sec2}

\subsection{Space Debris and Collision Avoidance}

\begin{figure}
\centering
\includegraphics[width=0.94\textwidth]{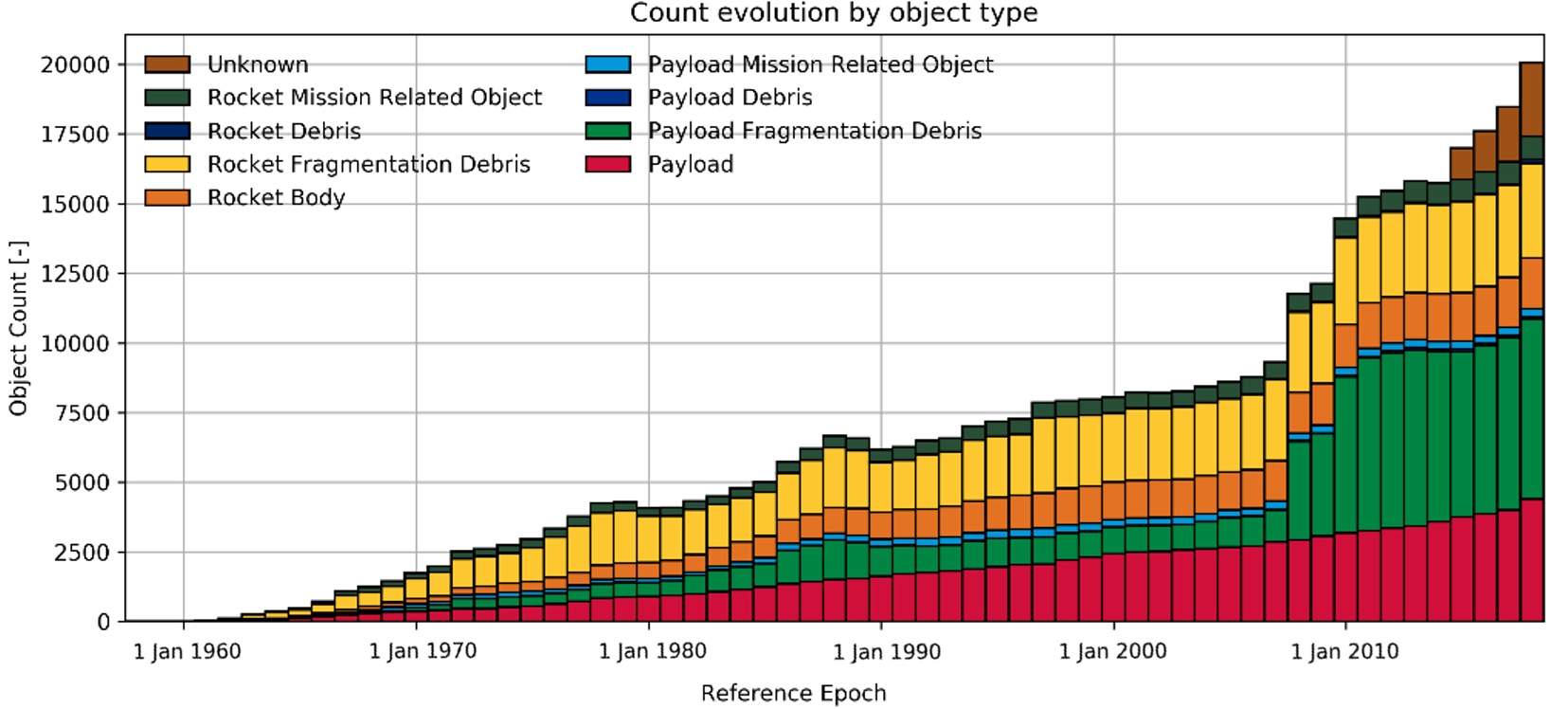}
\caption{The evolution in the number of space debris over the years~\cite{004}.}
\label{001}
\end{figure}

The escalating density of orbital debris -- well shown in \textsc{Figure}~\ref{001} -- poses a significant and growing threat to operational spacecraft, necessitating the implementation of robust collision avoidance strategies~\cite{003}. \\
Currently, a combination of on-ground and on-board solutions is employed to mitigate this risk. On-ground systems, primarily operated by organizations like the U.S. Space Force's 18th Space Control Squadron~\cite{005}, continuously track cataloged objects and predict potential close approaches (conjunctions) between debris and active satellites. Based on these predictions, satellite operators receive Conjunction Data Messages (CDMs) and can maneuver their spacecraft to avoid collisions. However, the accuracy of these predictions is inherently limited by uncertainties in debris tracking, orbital propagation models, atmospheric interference and the size of detectable objects~\cite{006}. To complement ground-based efforts, on-board collision avoidance systems are increasingly being considered by the scientific community~\cite{007}. These systems can independently assess collision risks using on-board sensors or by processing tracking data, enabling faster reaction times and greater autonomy in executing maneuvers to avoid a possible collision, particularly for uncatalogued or newly fragmented debris. The advantages of incorporating on-board solutions are manifold, including reduced reliance on ground segment communication latency and the ability to react to unforeseen conjunctions more swiftly. 
 \\
Notably, the implementation of ML and statistical approaches within on-board systems holds significant promise for enhancing collision avoidance capabilities~\cite{008}. ML algorithms can learn complex patterns from historical conjunction data and sensor readings to improve the accuracy of risk assessments, while statistical methods can provide probabilistic estimates of collision risk, allowing for more informed decision-making regarding maneuver execution. This shift towards intelligent on-board systems offers the potential for more proactive and autonomous debris mitigation, ultimately contributing to the long-term sustainability of space operations.

\subsection{Event-Based Cameras in Computer Vision}

\begin{figure}
\centering
\includegraphics[width=0.60\textwidth]{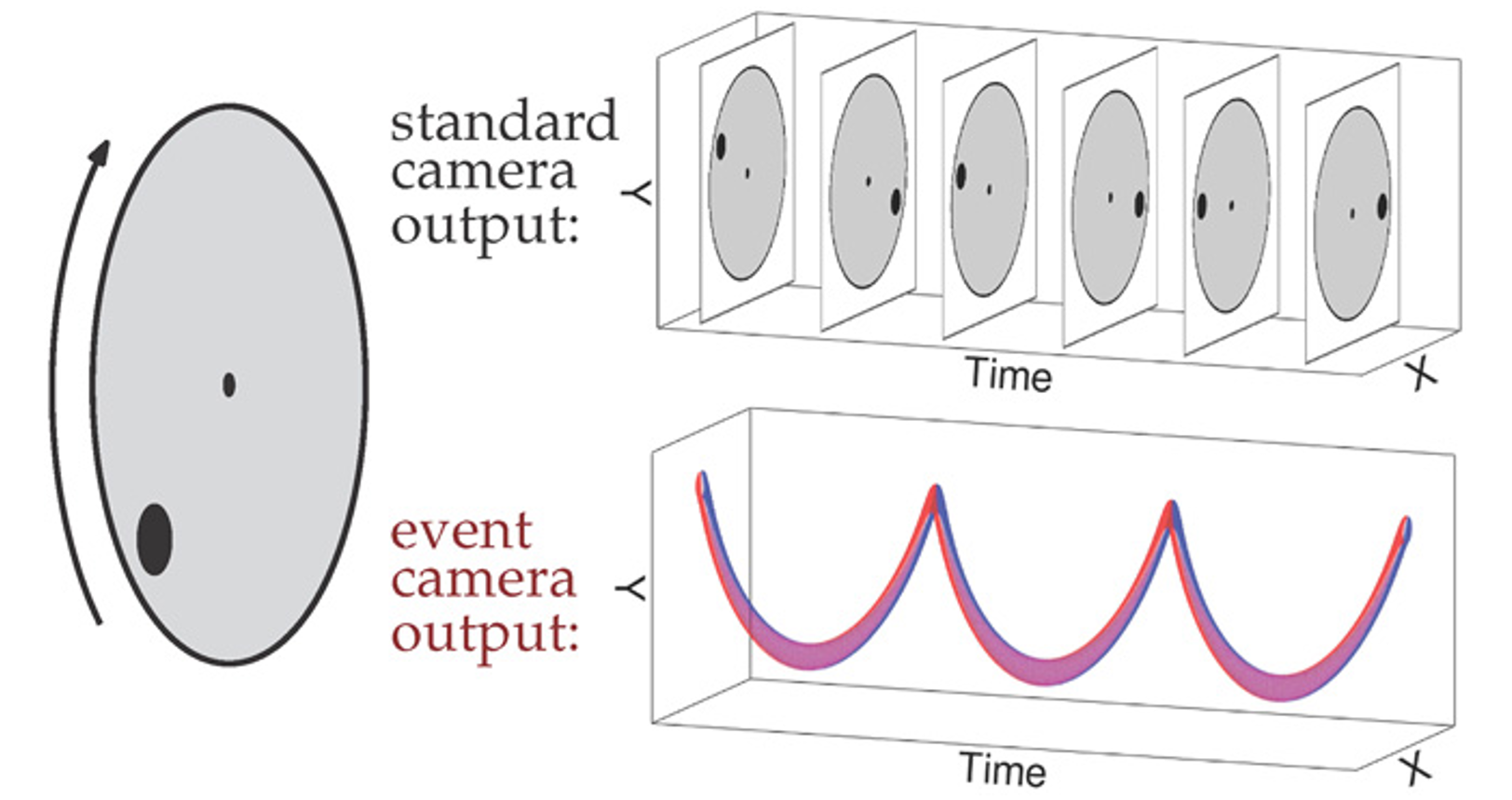}
\caption{A standard frame-based camera in comparison with an event-based camera~\cite{010}. The event-based camera activates its array of pixels only in case a change in light is detected, getting to lower latencies and higher temporal resolution.}
\label{002}
\end{figure}

Event-based cameras have revolutionized computer vision applications in robotics and autonomous systems~\cite{009} by addressing key limitations of traditional frame-based sensors through their asynchronous, event-driven architecture~\cite{010}. Unlike conventional cameras constrained by fixed frame rates, these devices capture intensity changes at microsecond resolution, enabling real-time tracking of high-speed motions (refer to \textsc{Figure}~\ref{002}). For instance, in aggressive drone racing scenarios, they excel under extreme apparent scene motion and proximity to obstacles, providing precise state estimation critical for navigation~\cite{011}. Their vast dynamic range (exceeding $140\,\mbox{dB}$) also allows robust operation across varying lighting conditions, outperforming standard sensors limited by fixed exposure times. Furthermore, their event-driven nature reduces data redundancy and power consumption, making them ideal for resource-limited platforms and autonomous systems. ML has further amplified their potential: optimized algorithms have improved tasks such as optical flow estimation and depth prediction~\cite{012}. Direct tracking methods using photometric 3D maps, which model event streams probabilistically, achieve superior localization accuracy in dynamic environments compared to frame-based approaches. These advances underscore the transformative role of event-based cameras in advancing computer vision for real-time, high-dynamic-range applications.

\subsection{ML Approaches for Space Object Recognition}

In modern collision avoidance for both on-board and ground-based debris detection systems, the use of ML techniques is envisioned. On-board detection systems will employ an array of sensors -- such as optical cameras, radars, and LIDAR -- to continuously scan the local space environment, thereby enabling satellites to autonomously identify, classify, and track nearby objects in real time. When these sensors are coupled with ML algorithms, the on-board systems can quickly analyze sensor data to predict potential collision trajectories and initiate evasive maneuvers without waiting for ground intervention~\cite{012}. Complementarily, ground-based detection systems harness data from extensive radar networks and telescopic observations to build and maintain detailed catalogs of orbital debris. ML models are then applied to these large datasets to refine orbital predictions, filter out noise, and assess collision probabilities with enhanced accuracy. \\
Recent research has demonstrated that supervised learning methods, such as deep neural networks and support vector machines, can effectively process historical tracking data to forecast debris trajectories and classify objects based on their motion characteristics~\cite{013}. Moreover, probabilistic approaches employing Bayesian ML have been used to incorporate uncertainty into collision risk estimates, offering more robust and reliable predictive models~\cite{014}. Reinforcement learning has also emerged as a promising tool for optimizing collision avoidance maneuvers. By learning from simulated space encounters through reward-based feedback, these algorithms iteratively improve decision-making policies that govern both evasive actions and subsequent orbit adjustments~\cite{012}. This combination of real-time on-board autonomy and enhanced ground-based tracking represents a significant advancement over traditional methods that relied on manual monitoring and delayed responses. \\
Interdisciplinary efforts are further propelling the evolution of these systems. For example, initiatives such as ESA’s collision avoidance service and open-source projects like Kessler -- a dedicated ML library for spacecraft collision avoidance -- illustrate the potential of integrated approaches to mitigate the growing threat of space debris~\cite{015}\cite{016}. As satellite constellations expand and the orbital environment becomes increasingly congested, the synergy between advanced sensor technologies and sophisticated ML models is critical to ensuring the long-term sustainability of space operations. \\
Future advancements will encompass the integration of ground-based analysis with reactive on-board systems capable of autonomous decision-making and enhancing the capabilities of space assets. These developments are driving the authors of this paper to explore an innovative on-board detection method, aiming to eventually develop Commercial Off-The-Shelf (COTS) components that can be implemented in every satellite in orbit.

\section{Methodology} \label{sec3}

\subsection{System Architecture}

\begin{figure}
\centering
\includegraphics[width=1.00\textwidth]{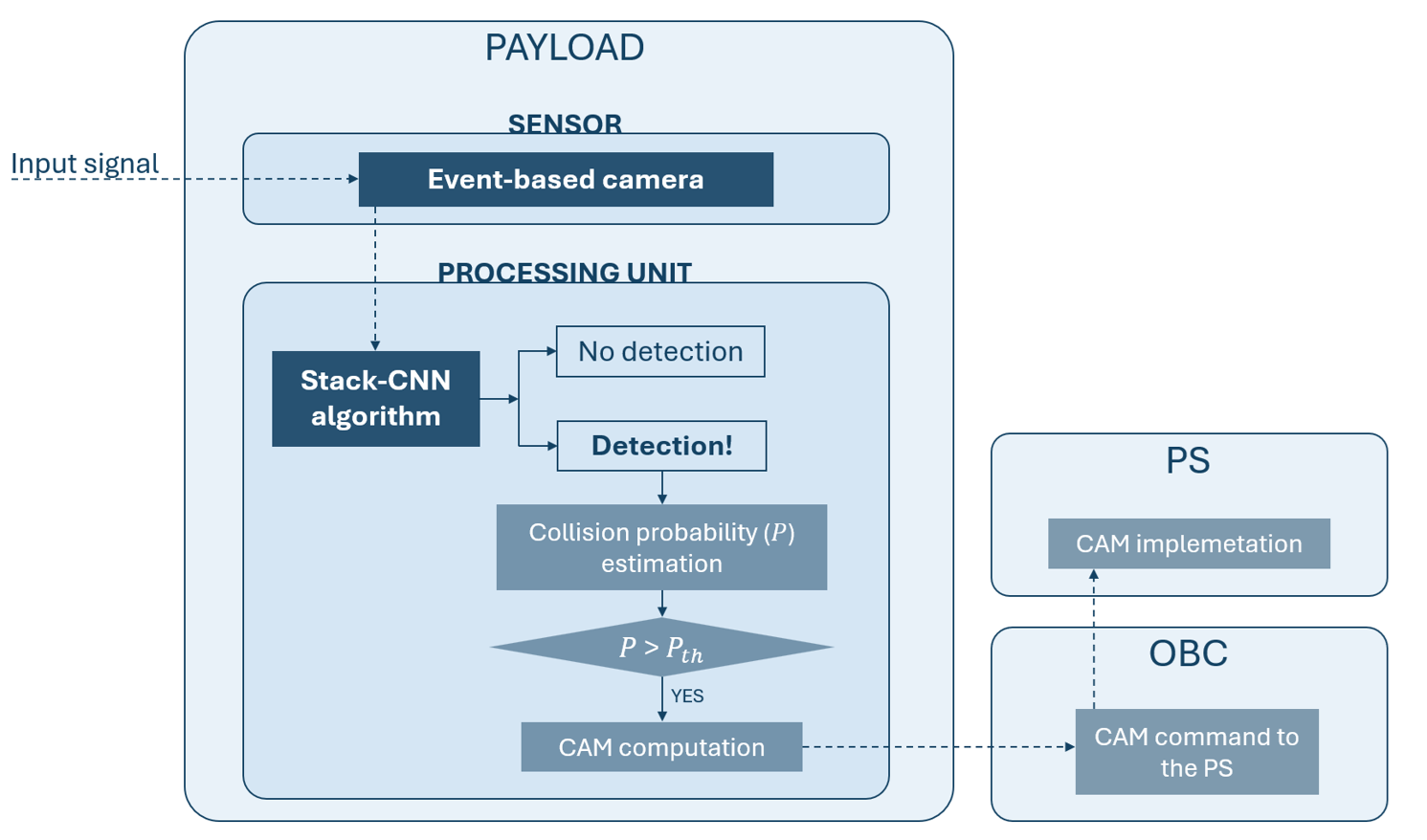}
\caption{Operational sequence of actions from the event-based camera input to the Stack-CNN algorithm processing, up to the CAM implementation. The collision probability estimation and CAM computation modules have not been addressed in this study.}
\label{003}
\end{figure}

The system proposed in this study consists of two main parts: a sensor, identified in an event-based camera, and a Processing Unit (PU) on which the Stack-CNN deep learning algorithm is implemented to trigger the passage of any bright object. \\
The neuromorphic camera is designed to be in constant listening mode and sends a continuous stream of data to the PU. These are stored in a buffer where they pass through the Stack-CNN algorithm, which in real time determines the possible presence of a moving light signal in the detector's Field of View (FOV). In case of detection, a tracking algorithm (not covered in this study), also implemented on the PU, will analyze the morphology of the light signal and its evolution through the event-based camera dataflow, defining as far as possible its trajectory within a certain error margin. Based on this, the probability of collision with the detected object (P) is computed. If the latter is smaller than a threshold probability, the collision is discarded. If, on the other hand, the probability is higher than the threshold, a CAM must be carried out as soon as possible. The CAM, then, is first computed by the payload's PU itself, and reported to the On-Board Computer (OBC), which will then command it to the Propulsion System (PS). \\
The implementation of the maneuver by the PS will avoid a possible collision. \textsc{Figure}~\ref{003} schematically shows the high-level functioning of the proposed system, and its interaction with OBC and PS.

\subsection{Event-based Camera Operation} \label{sec3.2}

The typology of sensor proposed in this study offers a novel way of storing the visual scene. In fact, an Event camera can sample the light entering its FOV depending on the dynamics of the scene~\cite{010}, as opposed to a traditional camera which samples the scene at a fixed sampling rate, storing static scenes as well. Compared to a conventional frame camera, this innovative neuromorphic sensor asynchronously measures per-pixel brightness changes and outputs a stream of events that encode time, location and polarity of each detected change. The polarity of the event depends on the dynamics of the scene, which will produce an increase (positive polarity events) or a decrease (negative polarity events) in the brightness levels associated with each pixel. \\
Compared to traditional cameras, these neuromorphic sensors offer higher temporal resolution (up to $1\,\mu\mbox{s}$) and wider dynamic range (up to $140\,\mbox{dB}$), essential features for accurate and complete detections of light-moving objects. In addition, the low latency and the very small data rate of these novel cameras allow a more rapid detection of fast-moving SD on a collision course, permitting timely mitigation actions. Finally, the low power consumption ($10-400\,\mbox{mW}$) and the small mass associated with these devices allow to impact on space mission’s power and mass budgets as little as possible, from CubeSats to bigger satellites. \\
For all these reasons, the choice of the detector fell on this innovative and efficient type of neuromorphic sensor.

\subsection{Stack-CNN algorithm}

\begin{figure}
\centering
\includegraphics[width=0.96\textwidth]{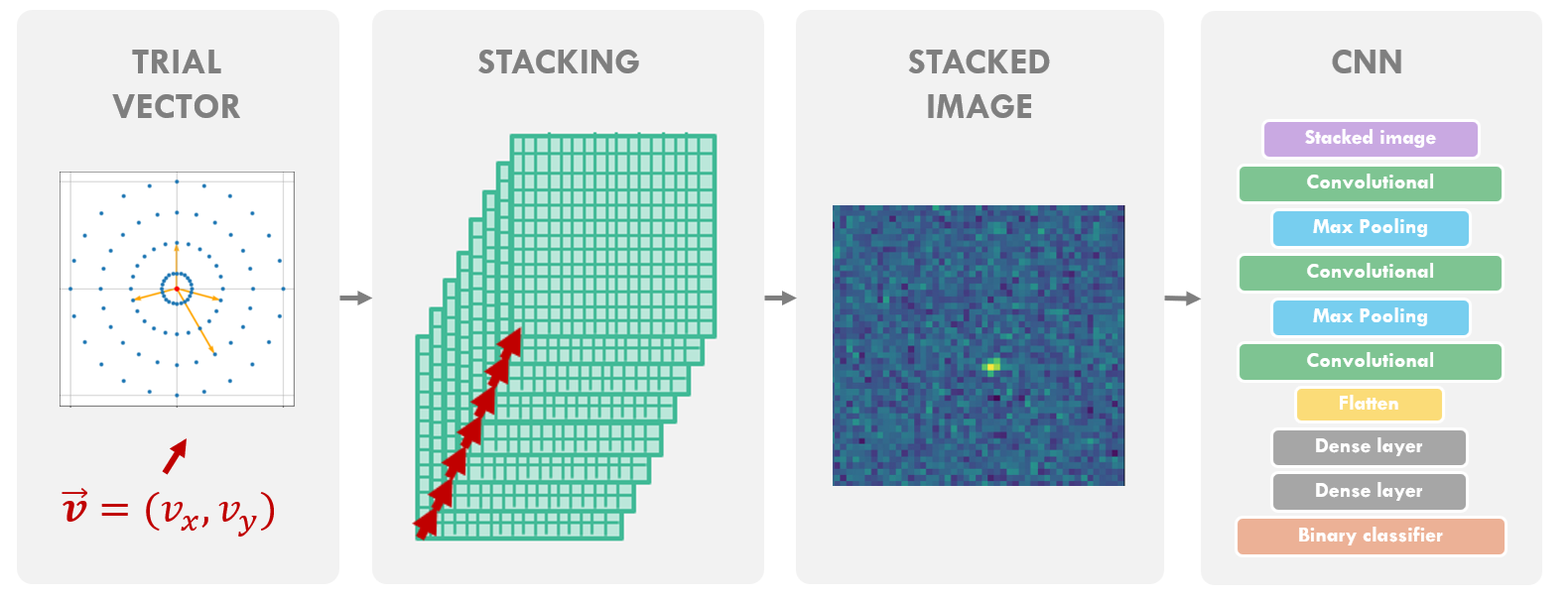}
\caption{Schematic working principle of the Stack-CNN method. After selecting a trial vector from the pool, the shift and add (Stacking) operations on consecutive frames are performed. The resulting Stacked image is then given to the CNN for the prediction. In the diagram, the shown Stacked image has a point-like shape, suggesting the selection of the correct trial vector.}
\label{004}
\end{figure}

Concerning the detection methodology, we chose to use a hybrid architecture that adopts a deep learning-based algorithm for feature extraction and detection of moving bright objects within the FOV of a detector. Specifically intended for the detection of SD or meteors~\cite{017}\cite{018}, this algorithm consists of two main steps: a first part called \textit{Stacking}, during which consecutive frames are shifted and summed according to a trial displacement vector, and a second part which involves the application of a CNN. Named \textit{Stack-CNN}, the algorithm can increase the SNR of the data due to the shifted sum of the frames. Indeed, in case the trial velocity vector according to which the Stacking procedure is performed corresponds to or is very near to the correct vector by which the signal is moving, the latter will be summed coherently producing a point-like bright area. Therefore, the SNR will be enhanced by a $\sqrt{n}$ factor, where $n$  is the number of frames used during the Stacking procedure. The following CNN solves a binary classification task and gives as output $'\,1\,'$ if the trial vector is very similar to the right one. Otherwise, a zero-output value will indicate either the use of an erroneous velocity vector or the total absence of a moving bright object. Refer to \textsc{Figure}~\ref{004} for an illustrative diagram of the Stack-CNN working principle. \\
As will be detailed in Section \ref{sec4}, the Stack-CNN algorithm has also been shown to be applicable to an asynchronous dataflow such as that produced by a neuromorphic sensor, making it a perfect candidate for use as part of a collision avoidance system. Due to its possibility to enhance the SNR, the algorithm can detect faint signals, pushing forward the observational limit distance and consequently giving more time to the CAM to be performed.

\section{Results} \label{sec4}

\subsection{Experimental Setup and Data}

\begin{figure}
\centering
\includegraphics[width=0.40\textwidth]{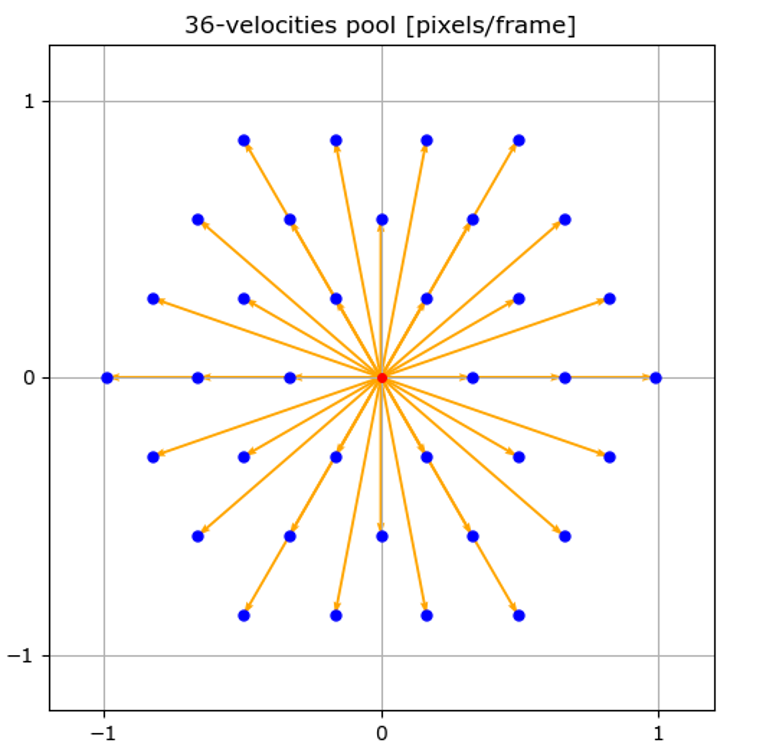}
\caption{Representation of the 36 trial velocity vectors used during the Stack-CNN performance test. This hexagonal configuration is optimized to cover the vector space as uniformly as possible, limiting to 1 the maximum pixels/frame displacement due to the necessity for the CNN to obtain a point-like bright area after the Stacking rather than a segment.}
\label{005}
\end{figure}

With the aim of simulating an in-orbit detection scenario involving a bright object passing through the FOV of an event-based camera, a public dataset collected using terrestrial event-based cameras pointed at the starry sky was used~\cite{019}. With over 8 hours of acquisition time, this Event-based Space Situational Awareness (EBSSA) dataset was the first publicly available dataset containing space imaging acquired by event-based cameras. \\
Using the Stack-CNN algorithm with a pool of $36$ trial velocity vectors optimized to be uniformly distributed in the two-dimensional vector space (see \textsc{Figure}~\ref{005}), the detection process was attempted on a 2-minutes data-taking session of event-based data containing 9 human-labelled bright objects moving in the FOV: one very bright and attributable to a launch vehicle stage (the SL-8 R/B~\cite{020}), tracked during its passage within the FOV of the camera, while the remaining eight associated with other kinds of space objects and presenting a wide range of event rates and SNRs.

\subsection{Pre-processing} \label{sec4.2}

To process the event-based data stream, a transformation of its content was first performed. We chose to use the \textit{simil-frame} representation~\cite{010}, accumulating the events over a fixed exposure time $\Delta t$. Afterwards, a spatial downsampling was performed aimed at reducing the original $240\times180$ spatial resolution in order to reduce not only the computational time associated with the Stacking procedure, but also the pixels/frame displacement of the bright moving objects (constrained to maximum 1 pixel/frame movements due to the Stack-CNN working principle). This spatial downsampling led to $80\times60$ frames.

\subsection{Performance and comparative analysis}

\begin{figure}
\centering
\includegraphics[width=0.94\textwidth]{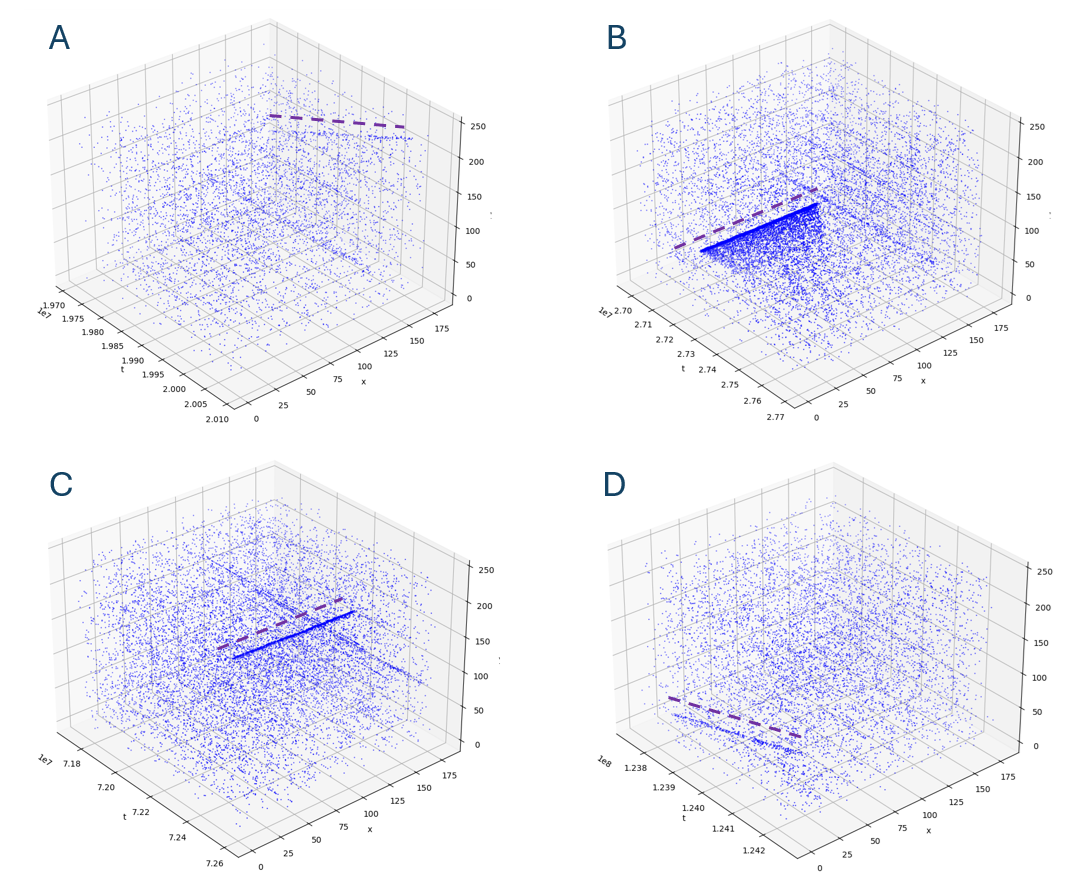}
\caption{Volumetric representation of the event-based data associated to four light-source events detected by the Stack-CNN algorithm in the SL-8 R/B data-taking session of the EBSSA dataset. The purple dashed line, parallel to the trace, helps its visualization. One very bright trace (B), two middle-bright traces (C, D) and one last very-low-SNR trace (A) are shown.}
\label{006}
\end{figure}

Once the pre-processing was completed, the run of the event-based data under the Stack-CNN algorithm was performed. This led to the detection of six out of nine light-moving objects, equaling the number of events detected by the algorithm proposed by the authors of the dataset themselves (see \textsc{Figure}~\ref{006} for the visual presentation of four of them). Attempting to avoiding the spatial downsampling, the run of the Stack-CNN algorithm was repeated, leading to the detection of all the nine bright traces. Thus, three faint objects, seen during the human-labelling procedure but not detected by the algorithm used by the authors of the dataset, are detectable by the Stack-CNN algorithm when the spatial downsampling of the frame is not performed. \\
This result demonstrates the soundness of the Stack-CNN algorithm used, underlining that it is able to process not only a traditional frame-based dataset~\cite{017}\cite{018}, but also an event-based dataset by means of an appropriate transformation of the latter. In addition, the Stack-CNN algorithm is proved to be more sensitive to fainter traces, finding more bright traces in the analyzed EBSSA data-taking session. In the next Section we will discuss the real-time potential associated with this methodology and its possible capabilities.

\section{Discussion} \label{sec5}

\subsection{Interpretation of Results}

The combination of an event-based camera with the Stack-CNN algorithm has been presented, demonstrating the possibility to successfully couple the two systems. \\
The advantages related to the use of a neuromorphic sensor, detailed in Section \ref{sec3.2}, can be summarized not only pointing out the efficiency of the camera for the purpose of detecting moving objects in general (storing only moving signals), but also underlining that it constitutes a perfect payload for SSA performed in space due to its suitable working parameters. \\
For what concerns the advantages of using the Stack-CNN detection algorithm, they are mainly linked to the possibility to enhance the SNR of the data through the Stacking procedure, detecting very faint signals and consequently increasing the observational limit distance. In this way, there is a larger time interval available to perform the CAM, increasing the capability of avoiding a collision with a space debris. Although the application of the Stack-CNN pipeline on the SL-8 R/B dataset shows similar detection results with respect to the algorithm used by the authors of the dataset~\cite{019} when performing the spatial downsampling of the datastream, the Stack-CNN algorithm is able to detect all the bright traces in the analyzed dataset (9 out of 9) when downsampling is avoided. Indeed, reducing the spatial resolution from $240\times 180$ to $80\times 60$ by grouping $3\times 3$ pixel blocks results in the dilution of the signal, making it fainter than it already is. Without applying the downsampling it is possible to enhance the Stack-CNN detection capabilities, but slowing down the analysis of the dataset. \\
This analysis proves the possibility for the algorithm to detect very faint traces due to the Stacking procedure. In addition, Stack-CNN is a detection method that can be able to operate in online mode, a crucial requirement for in-orbit collision avoidance purposes. 

\subsection{Limitations and Challenges}

\begin{figure}
\centering
\includegraphics[width=0.88\textwidth]{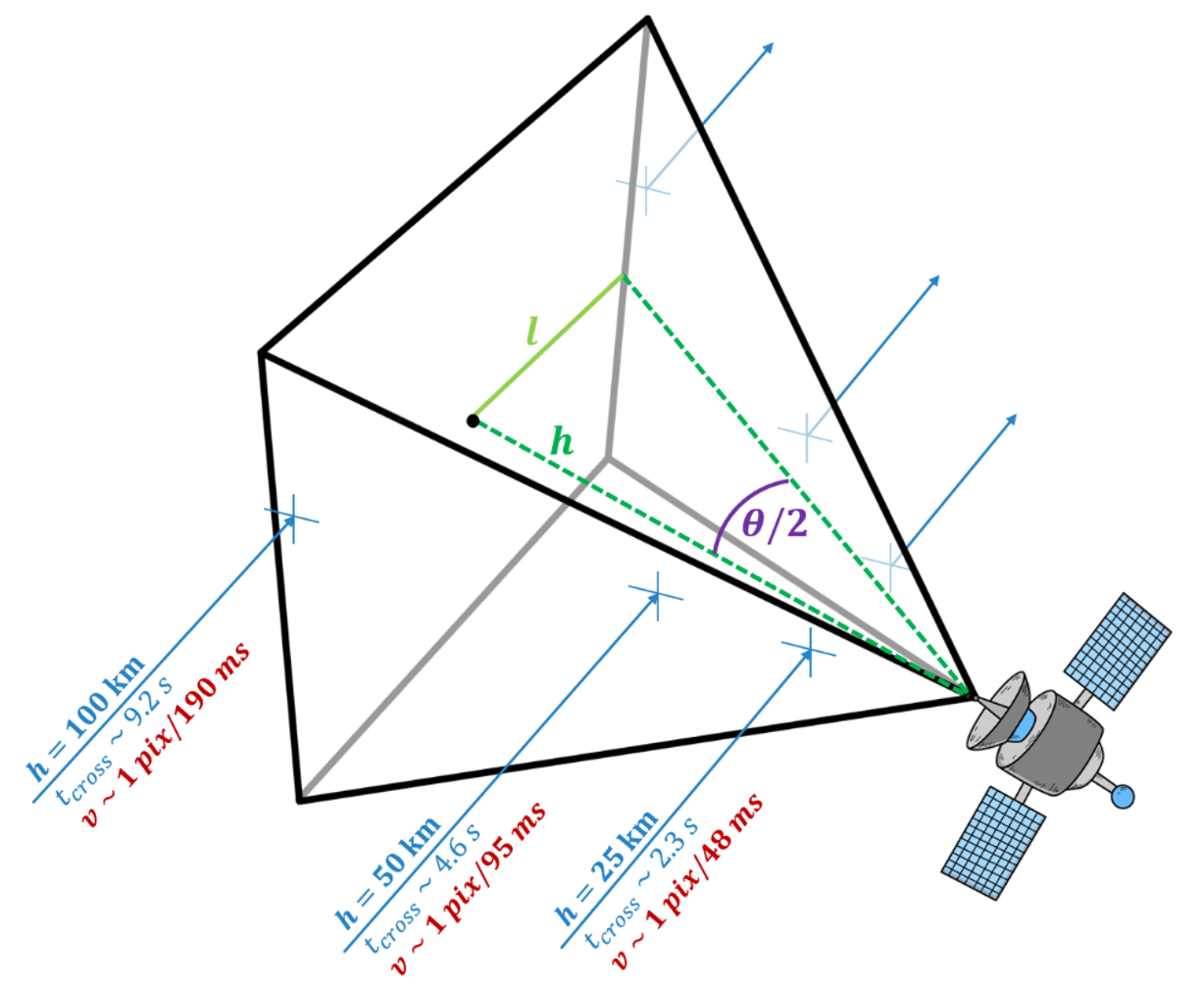}
\caption{The figure above shows 3 examples of space debris entering the FOV in the worst condition (velocity vector perpendicular to the FOV axis, producing the maximum displacement on the pixel matrix) at different distances from the sensor, highlighting the different displacement velocities on the pixel matrix (red notes). All the numbers are referred to a square FOV with $\vartheta = 40^{\circ}$, assuming a $48\times 48$ matrix and a space debris velocity equal to $7.5\,\mbox{km/s}$. Under these conditions, the trade-off choice for $\Delta t$ could be $80\,\mbox{ms}$, a time below which the Stack-CNN is most likely able to operate in real time, and able to see debris beyond $25\,\mbox{km}$ (even if the displacement exceeds 1 pixel/frame, Stack-CNN is still able to detect up to 1.5 pixels/frame displacements).}
\label{007}
\end{figure}

Although the combination of event-based camera and Stack-CNN seems to work, there are some critical issues depending on the task, i.e. the real-time detection of space debris entering the detector's FOV. \\
Firstly, for the algorithm to operate in real time, the frames analyzed should not be too time-resolved. As Stack-CNN operates on packets of consecutive frames to be shifted and summed, the higher is the temporal resolution, the greater will be the computational time. For this reason, the pre-processing step described in Section \ref{sec4.2} aims to group the events on a not too small sampling interval $\Delta t$. At the same time, the higher is the $\Delta t$, the more the benefits of the high temporal resolution camera will be lost, considering that time is a critical factor for a timely CAM.  \\
Secondly, the sampling time $\Delta t$ also determines the speed at which the signal moves frame to frame, resulting in faster displacements for higher $\Delta t$. In addition, the speed at which the debris moves in the pixel matrix is a function of the distance at which the debris enters the FOV (faster if it enters close to the detector, slower otherwise, see \textsc{Figure}~\ref{007}). However, since Stack-CNN is constrained to have a maximum speed of 1 pixel/frame to work properly, depending on the $\Delta t$ it is possible to define a distance below which the debris would not be detected because too fast. The higher is the sampling time $\Delta t$, the deeper is this threshold. For this reason, the choice of $\Delta t$ is critical, and should be a trade-off choice -- fixed all the other parameters involved in the problem -- to allow both the real-time operation of the algorithm and the observation of the greatest amount of debris by the Stack-CNN detection algorithm, maximizing the collision avoidance possibilities of the system. 

\subsection{Future Directions}

Future experimental studies will aim to quantify the detection sensitivity of an event-based camera in relation to its FOV, once coupled with the Stack-CNN method. This will make it possible to estimate the time span available to the spacecraft to perform a CAM, which gets smaller and smaller as the size of the SD decreases.  \\
In addition, an evolution of the approach for the detection task is currently being developed. Avoiding the explicit use of the Stacking procedure, it is characterized by a very low computational time, showing promising preliminary results in terms of detection capabilities. \\
Furthermore, to implement a completely neuromorphic pipeline, from the sensor (event-based camera) to the detection algorithm and the processor on which it is implemented, future studies will aim to convert the CNN of the algorithm into a Spiking-CNN, allowing it to be implemented on a neuromorphic chip. This will permit the detection system to be extremely efficient, minimizing the power consumption.

\section{Conclusions} \label{sec6}

Targeting a space-based collision avoidance system, this study analyses some preliminary aspects of combining an event-based camera -- a neuromorphic sensor particularly suited for use in space environments -- with the Stack-CNN online detection method -- a powerful deep learning algorithm capable of seeing extremely faint moving objects. The latter was tested on an EBSSA dataset and, thanks to its ability to increase the SNR of the acquired scene, was able to detect more faint traces compared to the algorithm used by the authors of the dataset. Due to this ability, the Stack-CNN algorithm might be able to anticipate the detection of a debris on a collision course, providing a larger time interval to perform a CAM. Future studies will aim to determine more precisely the sensitivity of the neuromorphic sensor when coupled with the Stack-CNN method, and the available time interval that this system would provide to operate a CAM in case of SD on a collision course. \\
Due to the high efficiency characteristics of the neuromorphic sensor, which is extremely well-suited for space application, and due to the excellent detection capabilities of the Stack-CNN algorithm, such a system could be extremely helpful in the future for in-space STM and SSA.


\begin{thebibliography}{999}

\bibitem{001}
\textsc{N. L. Johnson} (2010), “Orbital debris: the growing threat to space operations'', NASA.

\bibitem{002}
\textsc{United Nations} (1999), “Technical Report on Space Debris'', United Nations Publication, New York.

\bibitem{003}
\textsc{ESA}, “Mitigating space debris generation'', \url{https://www.esa.int/Space_Safety/Space_Debris/Mitigating_space_debris_generation}. Accessed on 5 April 2025. 

\bibitem{004}
\textsc{ESA}, “About space debris'', \url{https://www.esa.int/Space_Safety/Space_Debris/About_space_debris}. Accessed on 5 April 2025. 

\bibitem{005}
\textsc{US Space Force}, “18th Space Defense Squadron'', \url{https://www.spaceforce.mil/About-Us/Fact-Sheets/Fact-Sheet-Display/Article/3740012/18th-space-defense-squadron}. Accessed on 5 April 2025. 

\bibitem{006}
\textsc{T. Flohrer, J. Peltonen, A. J. Kramer, T. Eronen, et al.} (2005),
“Space-Based Optical Observations of Space Debris'', \textit{4th European Conference on Space Debris}, Darmstadt, Germany.

\bibitem{007}
\textsc{A. Menicucci, G. Drolshagen, J. Kuitunen, Y. Butenko, et al.} (2013), “In-Flight and Post-Flight Impact Data Analysis from DEBIE2 (Debris In-Orbit Evaluator) on Board of ISS'', \textit{6th European Conference on Space Debris}, Darmstadt, Germany.

\bibitem{008}
\textsc{K. Tsaprailis, G. Choumos, V. Lappas, C. Kontoes} (2024), “Survey Mode: A Review of Machine Learning in Resident Space Object Detection and Characterization'', \textit{AIAA SciTech Forum}, Orlando, USA.

\bibitem{009}
\textsc{S. Tenzin, A. Rassau, D. Chai} (2024), “Application of Event Cameras and Neuromorphic Computing to VSLAM: A Survey'', \textit{Biomimetics} \textbf{9}, p. 444.

\bibitem{010}
\textsc{G. Gallego, T. Delbruck, G. Orchard, C. Bartolozzi, et al.} (2022), “Event-Based Vision: A Survey'', \textit{IEEE Transactions on Pattern Analysis and Machine Intelligence} \textbf{44}, pp. 154-180.

\bibitem{011}
\textsc{J. Delmerico, S. Mintchev, A. Giusti, B. Gromov, et al.} (2019), “The current state and future outlook of rescue robotics'', \textit{Journal of Field Robotics} \textbf{2019}, pp. 1-21.

\bibitem{012}
\textsc{N. Bourriez, A. Loizeau, A. F. Abdin} (2023), “Spacecraft Autonomous Decision-Planning for Collision Avoidance: a Reinforcement Learning Approach'', \textit{74th International Astronautical Congress (IAC)}, Baku, Azerbaijan.

\bibitem{013}
\textsc{S. Lice} (2024), “Machine Learning Algorithms for Automated Space Debris Tracking and Collision Avoidance in Low Earth Orbit'', \textit{Journal of Astrophysics \& Aerospace Technology} \textbf{12}.

\bibitem{014}
\textsc{J. S. Catulo, C. Soares, M. Guimarães} (2023), “Predicting the Probability of Collision of a Satellite with Space Debris: A Bayesian Machine Learning Approach'', \url{https://arxiv.org/abs/2311.10633}.

\bibitem{015}
\textsc{ESA}, “Automating collision avoidance'', \url{https://www.esa.int/Space_Safety/Space_Debris/Automating_collision_avoidance}. Accessed on 30 March 2025. 

\bibitem{016}
\textsc{Kessler repository}, \url{https://github.com/kesslerlib/kessler}. 

\bibitem{017}
\textsc{A. Montanaro, T. Ebisuzaki, M. E. Bertaina} (2022), “Stack-CNN algorithm: A new approach for the detection of space objects'', \textit{Journal of Space Safety Engineering} \textbf{9}, pp. 72-82.

\bibitem{018}
\textsc{L. Olivi, A. Montanaro, M. E. Bertaina, A. G. Coretti et al.} (2024), “Refined STACK-CNN for Meteor and Space Debris Detection in Highly Variable Backgrounds'', \textit{IEEE Journal of Selected Topics in Applied Earth Observations and Remote Sensing} \textbf{17}, pp. 10432 - 10453.

\bibitem{019}
\textsc{S. Afshar, A. P. Nicholson, A van Schaik, G. Cohen} (2020), “Event-Based Object Detection and Tracking for Space Situational Awareness'', \textit{IEEE Sensors Journal} \textbf{20}, pp. 15117-15132.

\bibitem{020}
\textsc{A. Aggarwal}, “SL-8 R/B satellite details 1968-040b norad 3230'', \url{https://www.n2yo.com/satellite/?s=3230}. Accessed on 30 March 2025.


\end{thebibliography}
\end{document}